\documentclass{article}

\usepackage{arxiv}

\usepackage[utf8]{inputenc} 
\usepackage[T1]{fontenc}    
\usepackage{hyperref}       
\usepackage{url}            
\usepackage{booktabs}       
\usepackage{amsfonts}       
\usepackage{nicefrac}       
\usepackage{microtype}      
\usepackage{lipsum}
\usepackage{graphicx}
\graphicspath{ {./images/} }
\usepackage{listings}
\usepackage{color} 
\definecolor{codegreen}{rgb}{0,0.6,0}
\definecolor{codegray}{rgb}{0.5,0.5,0.5}
\definecolor{codepurple}{rgb}{0.58,0,0.82}
\definecolor{backcolour}{rgb}{0.95,0.95,0.92}
\usepackage{amsmath}  
\usepackage{amssymb}  
\usepackage{tabularx} 
\usepackage{enumitem}

\lstdefinestyle{mystyle}{
	backgroundcolor=\color{backcolour},   
	commentstyle=\color{codegreen},
	keywordstyle=\color{magenta},
	numberstyle=\tiny\color{codegray},
	stringstyle=\color{codepurple},
	basicstyle=\ttfamily\footnotesize,
	breakatwhitespace=false,         
	breaklines=true,                 
	captionpos=b,                    
	keepspaces=true,                 
	numbers=left,                    
	numbersep=5pt,                  
	showspaces=false,                
	showstringspaces=false,
	showtabs=false,                  
	tabsize=2
}

\lstdefinelanguage{json}{
	basicstyle=\normalfont\ttfamily,
	numbers=left,
	numberstyle=\scriptsize,
	stepnumber=1,
	numbersep=8pt,
	showstringspaces=false,
	breaklines=true,
	frame=lines,
	backgroundcolor=\color{backcolour},
	stringstyle=\color{codepurple},
	literate=
	*{0}{{{\color{codepurple}0}}}{1}
	{1}{{{\color{codepurple}1}}}{1}
	{2}{{{\color{codepurple}2}}}{1}
	{3}{{{\color{codepurple}3}}}{1}
	{4}{{{\color{codepurple}4}}}{1}
	{5}{{{\color{codepurple}5}}}{1}
	{6}{{{\color{codepurple}6}}}{1}
	{7}{{{\color{codepurple}7}}}{1}
	{8}{{{\color{codepurple}8}}}{1}
	{9}{{{\color{codepurple}9}}}{1}
	{:}{{{\color{codepurple}:}}}{1}
	{,}{{{\color{codepurple},}}}{1}
	{\{}{{{\color{codepurple}\{}}}{1}
	{\}}{{{\color{codepurple}\}}}}{1}
	{[}{{{\color{codepurple}[}}}{1}
	{]}{{{\color{codepurple}]}}}{1},
}

\title{Agentic AI Sustainability Assessment for Supply Chain Document Insights}

\author{ \href{https://orcid.org/0009-0008-7513-1255}{\includegraphics[scale=0.06]{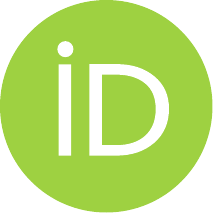}\hspace{1mm}Diego Gosmar} \\
	Head of AI, Tesisquare\\
	Voiceinteroperability.ai Initiative Member\\
	Linux Foundation AI \& Data\\
	\texttt{diego.gosmar@ieee.org} \\
	\And
	\href{https://orcid.org/0009-0007-0028-7573}{\includegraphics[scale=0.06]{orcid.pdf}\hspace{1mm}Anna Chiara Pallotta} \\
	Functional Analyst, Tesisquare\\
	MSc in Engineering Management\\
	Polytechnic University of Turin\\
	\texttt{annachiara.pallotta@tesisquare.com} \\
	\And
	\href{https://orcid.org/0000-0002-0996-6739}{\includegraphics[scale=0.06]{orcid.pdf}\hspace{1mm}Giovanni Zenezini} \\
	Assistant Professor in PM and Supply Chain Management\\
	Polytechnic University of Turin\\
	\texttt{giovanni.zenezini@polito.it}\\
}

\begin{document}
	
\maketitle
	\begin{abstract}
		
		This paper presents a comprehensive sustainability assessment framework for document intelligence within supply chain operations, centered on agentic artificial intelligence (AI). We address the dual objective of improving automation efficiency while providing measurable environmental performance in document-intensive workflows. The research compares three scenarios: fully manual (human-only), AI-assisted (human-in-the-loop, HITL), and an advanced multi-agent agentic AI workflow leveraging parsers and verifiers. Empirical results show that AI-assisted HITL and agentic AI scenarios achieve reductions of up to 70--90\% in energy consumption, 90--97\% in carbon dioxide emissions, and 89--98\% in water usage compared to manual processes. Notably, full agentic configurations---combining advanced reasoning (thinking mode) and multi-agent validation---achieve substantial sustainability gains over human-only approaches, even when resource usage increases slightly versus simpler AI-assisted solutions. The framework integrates performance, energy, and emission indicators into a unified ESG-oriented methodology for assessing and governing AI-enabled supply chain solutions. The paper includes a complete replicability use case demonstrating the methodology's application to real-world document extraction tasks.
	\end{abstract}
	
\section*{Structure of the Paper}

This paper is structured as follows. Section~\ref{sec:introduction} introduces the strategic role of agentic AI in supply chain document intelligence, framing the dual challenge of operational efficiency and environmental sustainability. Section~\ref{sec:foundations} establishes the theoretical foundations of intelligent and agentic systems, reviewing classical agent architectures, the Belief-Desire-Intention (BDI) framework, and the operational principles of Large Language Model (LLM)-based agents within Human-in-the-Loop (HITL) paradigms. Section~\ref{sec:llm_capabilities} examines advanced LLM agent capabilities, including orchestration strategies, tool utilization, and recent case studies in supply chain management applications. Section~\ref{sec:sustainability_framework} presents the sustainability assessment framework, detailing the environmental footprint of LLMs in terms of energy consumption, carbon emissions, and water usage, and introduces the operational scenarios evaluated in this research. Section~\ref{sec:comparative_analysis} provides a comprehensive comparative analysis of three document processing workflows: fully manual (human-only), AI-assisted (HITL), and advanced agentic AI with parser and verifier agents, quantifying reductions in energy, CO$_2$, and water usage. 

Section~\ref{sec:replicability} demonstrates the replicability of the agentic AI document extraction workflow through a concrete use case: extracting structured pricing data from a complex proforma invoice using Gemini 2.5 Flash with extended thinking capability, with complete token accounting, energy consumption analysis, and public GitHub implementation for community validation and benchmarking.
Section~\ref{sec:strategic_relevance} discusses the strategic relevance and governance implications, emphasizing the eco-efficiency trade-offs and organizational benefits of agentic automation. Section~\ref{sec:system_resilience} addresses system resilience and security considerations in multi-agent architectures. Section~\ref{sec:future_directions} outlines future research directions, including life-cycle assessments, dynamic context engineering, and lightweight autonomous agents. Finally, Section~\ref{sec:conclusions} synthesizes the key findings and reinforces the strategic imperative of integrating sustainability metrics into AI governance frameworks for supply chain transformation.
	
\section{Introduction: Agentic Transformation of Global Supply Chains}
\label{sec:introduction}
The contemporary landscape of global supply chain networks is defined by profound complexity, intensive digitalization, and exponential data growth, evolving into data-centric ecosystems that generate millions of documents daily, encompassing purchase orders, invoices, delivery notes, and critical Environmental, Social, and Governance (ESG) reports \cite{rao2022greenai, coresite2024sustainability, strubell2019energy}. This environment necessitates a strategic shift from poorly integrated, sequential operational models towards highly automated, resilient, and adaptive systems \cite{patterson2021carbon, sciencedirect2025_carbon}.
This research was developed through a collaboration between Tesisquare, an international software company specializing in supply chain collaboration \cite{tesisquare2025}, and the Polytechnic University of Turin's Department of Management Engineering \cite{polito2025managementengineering}.

The empirical case study and operational data were provided by Tesisquare's industrial implementation of AI-assisted document processing workflows, enabling a rigorous assessment of both operational efficiency and environmental sustainability metrics in real-world supply chain operations. The analysis encompassed the processing and extraction of nearly 80 different document types spanning across multiple supply chain stakeholders: suppliers (providing raw materials and components), manufacturers (transforming raw materials into finished goods), distributors and retailers (managing inventory and sales), and end customers (acquiring finished products). The document types included in the study primarily covered invoices, proforma invoices, purchase orders, advanced shipping notices (ASNs), shipment notes, and contractual agreements—representing the critical transactional backbone of multi-enterprise collaborative ecosystems. 

All documents were fully anonymized to protect sensitive business information, proprietary data, and commercial confidentiality, ensuring compliance with privacy regulations while preserving the empirical validity and generalizability of the findings. This heterogeneous and high-volume document landscape reflects real-world supply chain complexity and challenges, providing a robust empirical foundation for assessing the scalability, accuracy, and environmental impact of agentic AI solutions across diverse operational contexts.

Historically, the flow of information lagged behind the physical movement of materials, frequently resulting in operational bottlenecks, inventory imbalances, delayed deliveries, and inadequate traceability \cite{henderson2022systematicreportingenergycarbon}. Foundational technologies such as the Internet of Things (IoT), Big Data Analytics, and Cloud Computing have provided the infrastructure for real-time data sharing and intelligent decision-making; however, the manual processing of high-volume documents remains a critical constraint, associated with high error rates and time-intensive human intervention \cite{patterson2021carbon}.

\subsection{The Role of Artificial Intelligence and the need of Agentic Systems}

Artificial intelligence (AI) has become indispensable for leveraging the massive quantity of data generated across the supply chain. Moving beyond foundational predictive analytics based on historical trends, modern AI supports sophisticated functions including risk management, production planning, inventory optimization, and multi-criteria supplier selection that integrates explicit sustainability metrics \cite{Piccialli2025}.
The advent of Large Language Models (LLMs) represents a qualitative methodological transition, evolving from simple generative systems into \textbf{Agentic AI}: autonomous software entities characterized by advanced cognitive functions \cite{Piccialli2025}, and capable of interacting with each other by forming scalable Multi-Agent Systems (MAS) \cite{gosmar2024aimultiagentinteroperabilityextension}. These agents are capable of sophisticated reasoning, multi-step workflow planning, and dynamic action upon the external environment through integrated tool utilization, enabling their integration into complex enterprise processes. This capability is particularly transformative in document intelligence, where the agent shifts from basic Optical Character Recognition (OCR) to semantic interpretation, resolving textual ambiguities and structuring output data for enterprise resource planning (ERP) systems \cite{strubell2019energy}.
In this research, Agentic AI principles are applied to optimize document processing workflows, embedding cognitive automation within human-supervised operational loops rather than pursuing full autonomy.

\subsection{The Imperative of Green AI Governance}

Despite the demonstrable operational benefits derived from agentic systems, their widespread deployment introduces a significant challenge related to resource intensity \cite{patterson2021carbon, sciencedirect2025_carbon}. LLM-based agents, especially those relying on high-parameter foundation models, impose massive computational demands \cite{mit2025_generativeAI}, primarily requiring high-power Graphical Processing Units (GPUs) during the inference phase \cite{strubell2019energy}. This computational load substantially increases energy consumption, elevates the overall carbon footprint ($\text{CO}_{2}$) of the IT infrastructure \cite{rao2022greenai, strubell2019energy}, and drives up operational costs. The environmental impact of a single complex LLM query can be up to 50 times greater than a simple query due to complex reasoning processes \cite{henderson2022systematicreportingenergycarbon}.
For AI to fulfill its potential as a sustainable strategic lever—a central tenet of the Green AI movement \cite{rao2022greenai} and the evolution towards Industry 5.0/6.0—sustainability metrics must be systematically integrated into the system design and deployment architecture \cite{patterson2021carbon, henderson2022systematicreportingenergycarbon}. This necessitates the standardized quantification of environmental cost, specifically energy consumption ($\text{kWh}$), carbon dioxide emissions ($\text{kg}$), and water usage ($\text{L}$) \cite{irdo2025_openai, henderson2022systematicreportingenergycarbon}. The core objective, therefore, is to achieve highly efficient automation without compromising long-term ecological sustainability goals.

\section{Foundations of Intelligent and Agentic Systems with HITL}
\label{sec:foundations}
Understanding the paradigm shift from classical computational programs towards modern Large Language Model (LLM) agents necessitates a formal review of the foundational principles of intelligent systems, particularly regarding their internal architecture and operational cycles. Within the context of supply chain document intelligence, these principles are instantiated through AI agents that collaborate with human operators instead of replacing them, thereby enhancing operational resilience and ensuring high-quality outcomes.
A critical element of this collaborative paradigm is embodied in the Human-in-the-Loop (HITL) approach, which combines the strengths of data-driven machine learning models with the contextual expertise of human operators. As outlined by Sriraam Natarajan \textit{et al.} \cite{hitl}, this approach emphasizes that such systems are better understood as \textit{AI-in-the-Loop} (AI\textsuperscript{2}L) systems—where humans remain the primary agents controlling the system, with AI providing support rather than control. This distinction shifts the focus from mere AI performance to the interactive dynamics between human judgment and algorithmic assistance.
The integration of human expertise directly addresses several limitations intrinsic to fully automated models. HITL systems utilize human input to guide, validate, and correct algorithmic outputs, which is especially valuable in environments marked by high uncertainty and complex decision-making processes. This interaction not only improves the overall robustness and reliability of the system but also fosters greater transparency and interpretability, essential for building trust and ensuring accountability in AI applications \cite{hitl}.
Furthermore, maintaining humans as active participants supports continuous learning and adaptation. Through iterative human feedback, AI components can evolve, learning from new operational contexts and expertise, thereby creating a hybrid intelligence where the collective capabilities of humans and machines surpass what either can accomplish alone. This synergy is vital for mitigating biases and errors, contributing toward more ethical and equitable AI deployment.
In sum, the theoretical grounding of HITL systems underscores the importance of human agency as an essential element of system architecture. Recognizing humans as active, controlling participants rather than passive data sources represents a fundamental shift towards more trustworthy, effective, and comprehensive AI systems, particularly in high-stakes environments such as supply chain management and healthcare.

\subsection{Classical Intelligent Agents: Properties and Operational Cycles}

An intelligent agent is a system that perceives its environment through sensors and acts upon it using actuators. Intelligence emerges through four foundational properties: Autonomy (independent operation and adaptation), Reactivity (prompt response to environmental changes), Proactivity (initiative to pursue objectives), and Sociality (interaction and coordination with other entities, forming Multi-Agent Systems) \cite{gosmar2024conversationalaimultiagentinteroperability, gosmar2024aimultiagentinteroperabilityextension}.

Agent operation follows the Perception-Reasoning-Action (P-R-A) loop \cite{tang2025semanticintelligencebioinspiredcognitive,sapkota2025aiagentsvsagentic}, where the agent collects inputs, processes them via decision logic, and executes interventions. This is formalized as $f: P^{*}\rightarrow A$, mapping percept sequences ($P^{*}$) to actions ($A$). Agent architectures range from reactive (simple stimulus-response) to deliberative (symbolic reasoning) models. The Belief-Desire-Intention (BDI) framework represents a hybrid approach managing cognitive complexity through Beliefs (knowledge), Desires (objectives), and Intentions (committed plans) \cite{L_veill__2025}.

\subsection{The LLM Core and its Agentic Capabilities}

Large Language Models fundamentally enhanced agent reasoning through Transformer architectures and massive pre-training, achieving emergent capabilities in complex reasoning and contextual understanding \cite{bender2021dangers}. Enterprise deployment requires adaptation via Supervised Fine-Tuning (SFT) and Reinforcement Learning from Human Feedback (RLHF). Parameter-Efficient Fine-Tuning (PEFT) methods like LoRA and DORA enable specialization by adjusting focused parameter subsets rather than billions of weights \cite{henderson2022systematicreportingenergycarbon}, making industrial deployment economically viable.

\subsection{The Think-Act-Learn Cycle and Memory Management}

LLM-powered agents refine the P-R-A loop into the Think-Act-Learn cycle: analyzing input and formulating plans (Think), executing actions via tools and APIs (Act), and integrating feedback into experiential memory (Learn). This aligns with multi-agent conversational frameworks enabling universal APIs for multimodal communication \cite{gosmar2024conversationalaimultiagentinteroperability}. Retrieval-Augmented Generation (RAG) enables dynamic querying of Long-Term Memory via Vector Databases, injecting relevant context into prompts (Short-Term Memory) to ensure actions reflect full operational history and institutional knowledge.

\section{Advanced LLM Agent Capabilities and Applied Research in SCM}
\label{sec:llm_capabilities}
The operational efficacy of AI-assisted systems in supply chain management (SCM) depends on the sophistication of their control mechanisms and their integration within human workflows.

\subsection{Orchestration and Advanced Prompting Strategies}

Prompt engineering serves as the deterministic interface to control the LLM agent's behavior. For industrial extraction tasks, the model temperature is typically set to zero to ensure deterministic, predictable output.
Sophisticated prompting strategies are utilized to elicit complex thought processes via Few-Shots and Dynamic Template-based prompt engineering. Chain-of-Thought (CoT) compels the LLM to articulate its reasoning step-by-step, vital for complex logical tasks. Self-Refinement Prompting involves the model performing an auto-critique of its draft output, iteratively enhancing quality. Decomposed Prompting systematically breaks down complex problems into smaller, sequential sub-problems. The primary mechanism enabling the "Act" phase is Tool and Function Calling, which transforms the LLM into an active component capable of performing actions via external APIs.
Emerging interoperability extensions now allow coordination of multiparty conversational contexts across multiple LLM agents, improving orchestration and turn-taking in distributed agentic environments \cite{gosmar2024aimultiagentinteroperabilityextension}.

\subsection{Recent Case Studies in Agentic Supply Chain Management}

Academic and industrial research confirms the diverse applicability of LLM agents across core SCM functions. Autonomous Negotiation agents demonstrated superior outcomes in simulated supply chain contract negotiations, driven by tailored RAG configurations that enhance contextual memory for achieving specific bargaining objectives \cite{Kirshner2024}. The InvAgent multi-agent framework \cite{quan2025invagentlargelanguagemodel} utilized LLMs in a multi-echelon supply chain simulation to determine optimal ordering quantities, demonstrating that LLMs could make high-quality inventory decisions using zero-shot reasoning enhanced by CoT techniques, offering superior transparency compared to traditional black-box Reinforcement Learning (RL) models. Empirical validation confirmed the industrial scalability of a hierarchical LLM agent architecture for complex supply chain planning in the SCPA Framework \cite{qi2025leveragingllmbasedagentsintelligent}, achieving a reduction in weekly data processing time of approximately $40\%$ and an increase in plan accuracy of $22\%$ in a real retail environment. Furthermore, LLM-based Smart Audit Systems \cite{yao2024smartauditempoweredllm} have been developed to analyze extensive textual audit data against historical non-conformities, leading to a reduction of audit time by approximately $24\%$. Similarly, prototypes detect potential disruption events and visualize supplier vulnerabilities using Knowledge Graphs (KG) \cite{Zhang2025}.
These studies collectively support a hybrid vision of automation, where AI agents act as cognitive collaborators to human experts—consistent with the human-in-the-loop paradigm explored in this research.

\section{Sustainability Assessment: Framework and AI-Assisted Scenario}
\label{sec:sustainability_framework}
This section provides the environmental and energy assessment underpinning the comparative analysis between the traditional human-only document processing workflow and the AI-assisted (HITL) automation scenario implemented in Tesisquare's experimental context. A final, highly relevant aspect of the critical analysis concerning the LLM utilized by the developed AI agent involves environmental sustainability. Recent literature highlights that the operational and organizational benefits derived from using LLMs must be balanced with their energy and environmental impact \cite{rao2022greenai}.

\subsection{The Environmental Footprint of Large Language Models}

The training, inference, and maintenance of these models often require energy-intensive infrastructure, with daily electricity consumption potentially reaching hundreds of megawatt-hours. Specialized data center cooling systems are essential to maintain the stability and performance of the thousands of computing units that power these models \cite{chen2025electricitydemandgridimpacts}.
Data centers that trainand host LLMs require complex cooling systems, necessary to dissipate the heat produced by the operational units. Especially when working in parallel for extended periods, these units produce enormous quantities of heat that must beeliminated to avoid malfunctions and maintain operational stability. For this reason, servers are often cooled with airconditioning or water systems, which entails a consistent use of water resources.

Further researches have introduced the concept of \textbf{Water Usage Effectiveness (WUE)} \cite{li2025makingaithirstyuncovering,shumba2024waterefficiencydatasetafrican}, an indicator that measures the volume of water consumed per $\text{kWh}$ of energy used by a data center. The WUE is linked to three different scopes of water consumption: On-site (Scope 1) water produced by evaporative towers to cool the data centers, Off-site (Scope 2) water necessary to produce the electricity that powers the data centers, and Embodied water (Scope 3) water used to manufacture and ensure the proper functioning of the hardware.

Experimental results show that the WUE (Water Usage Effectiveness) ranges from $0.18\,\text{liters/kWh}$ to $0.30\,\text{liters/kWh}$ for efficient infrastructures, and up to $1\,\text{liter/kWh}$ in less optimized data centers~\cite{google2021,masanet2020carbon}. This implies that the execution of an AI agent can result in water consumption of approximately $0.3$ to $0.5$ liters for every kWh of energy used. Furthermore, while training is the phase most analyzed for environmental impact, inference today represents the main issue, constituting about 90\% of the energy lifecycle of an LLM~\cite{schwartz2020green,patterson2021carbon}. This means that, even after the training or inference phase, maintaining a model in operation involves significant ongoing energy and water consumption.

\subsection{LLM-Agent Scenario: Data Center and Initial Consumption Analysis}

Translating these data into the corporate project provides a macroscopic analysis of the consumption related to the developed agent. The Google Gemini Flash model utilized is designed to be relatively lightweight, meaning the consumption per single elaboration is much lower than larger models like GPT-3 or BLOOM. However, the overall impact depends critically on the number of documents processed and the type of cloud infrastructure used.

Tesisquare tested the services offered by Google Cloud, which declared an average annual Power Usage Effectiveness (PUE) of 1.09 for its global fleet of data centers in 2024 \cite{google2024pue}. This means that for every 1 kWh used for computing, only an additional 0.09 kWh is consumed for overhead such as cooling. This reflects an energy efficiency approximately 84\% higher compared to the industry average PUE of 1.56 \cite{uptime2024pue}.

To estimate the total resource utilization, we propose a basic assumption of an average energy consumption of 0.50~Wh per processed document. This is consistent with the use of a lightweight model like Gemini Flash for an elaboration flow where only a single call per document is required, directly returning the requested JSON output without subsequent steps of reformatting or validation. Considering that literature suggests a model like Gemini consumes about 0.24~Wh for each medium-difficulty prompt elaboration \cite{google2025geminienergy}, this 0.50~Wh hypothesis is reliable given the prompt's complexity.

Applying the PUE of 1.09 results in a total energy requirement of 0.000545~kWh per document. Applying the efficient WUE (Water Usage Effectiveness) range of 0.18 to 0.30~L/kWh \cite{industry2024wue} leads to an estimated water consumption of 0.09 to 0.15~ml per document. Based on a daily processing volume of 5,000 documents, the consumption for the LLM agent scenario is approximately 2.725~kWh of energy and 0.45 to 0.75 liters of water per day. The trend in energy and water consumption grows proportionally with the number of processed documents, remaining dependent on the efficiency of the underlying data center infrastructure.

Moreover, incorporating Google's environmental impact estimates for AI inference, each prompt generates approximately 0.03 grams of CO$_2$ equivalent emissions \cite{google2025geminienergy}. Thus, for 5,000 documents daily, the total CO$_2$ emissions amount to about 150 grams per day, or 0.15 kilograms, demonstrating a relatively low carbon footprint attributable to the highly efficient, clean energy-powered data centers.

\section{Comparative Analysis: Human-Only vs AI-Assisted Workflow}
\label{sec:comparative_analysis}
The comparative analysis evaluates the operational efficiency and sustainability trade-offs between the manual (human-only) document management process and the AI-assisted (human-in-the-loop) workflow developed within Tesisquare's industrial use case.
Literature observations \cite{smith2024manvmachine} suggest that while language models imply significant energy and water consumption, they can contribute to reducing activities that would otherwise have a much more impactful environmental consequence. The overall environmental balance must therefore consider the savings generated by the transition from manual, paper-based management to automated digital management. The automation reduces the need for manual interventions, time, human resources, and the circulation of physical paper material, thus lowering environmental impacts connected to production, transport, and archiving.

\subsection{Methodology of the Comparative Evaluation}

To perform a robust comparative analysis, both the computational consumption of the data center and the impact associated with the operator's workstation must be considered, since even with the AI agent, a set of activities (file retrieval, call initiation, quick output verification) requires human intervention.

\subsubsection{Baseline Operational Assumptions}

In this analysis, a robust and conservative estimation approach has been applied to account for variability and uncertainties in operational parameters, operator productivity, and infrastructure efficiency. The reported ranges thus reflect prudent buffers ensuring that observed environmental benefits of AI assistance are realistically achievable across diverse real-world operational contexts.

\begin{enumerate}[label=\textbf{\arabic*}., itemsep=1.5ex]
	\item \textbf{Shift Configuration:} 8-hour workday with 7 productive hours, including a 15\% buffer for absences and inactivity.
	\item \textbf{Operational Scale:} 5,000 documents processed per day, representative of medium-to-large enterprise document workflows.
	\item \textbf{Workstation Consumption:} A typical laptop workstation consumes an average of 60 Wh/h \cite{energystar2025laptop}, equating to $\mathbf{0.48 \text{ kWh}/\text{day}}$ (8 hours). This is multiplied by the number of required operators.
	\item \textbf{Italian Emission Factor:} We use the Italian national average emission factor \cite{ispra2024efficiency} of $\mathbf{288 \text{ gCO}_2/\text{kWh}}$.
\end{enumerate}

\subsubsection{Processing Time and Throughput Assumptions}

\textbf{Manual (Human-Only) Scenario:}
\begin{itemize}
	\item Average processing time per document: 5 to 30 minutes
	\item Estimated productivity per operator: 14 to 84 documents/day
	\item Number of operators required: \textbf{70 to 400 operators}
\end{itemize}

\textbf{AI-Assisted (Human-in-the-Loop) Scenario:}
\begin{itemize}
	\item Average HITL time per document: 30 to 120 seconds (0.5 to 2 minutes)
	\item Estimated throughput per operator: 210 to 840 documents/day
	\item Number of operators required: \textbf{7 to 28 operators}
\end{itemize}

The comparison between the fully manual (human-only) workflow and the AI-assisted (human-in-the-loop) scenario reveals a significant sustainability gap. By introducing partial automation through the AI agent, the overall daily energy requirement is reduced from approximately 36.3--194.7 kWh to just 6.1--16.2 kWh. This represents an energy reduction between 70\% and 90\% compared to the traditional process. Applying the same emission factor (288 gCO$_2$/kWh) translates into a decrease in CO$_2$ emissions from roughly 10.5--56.1 kg/day down to 1.8--4.7 kg/day. Similarly, water usage drops from 35.1--58.4 L/day in the human-only case to 1.1--4.9 L/day in the AI-assisted scenario. Figure~\ref{fig:comparison} provides a comparative visualization of the maximum sustainability metrics across both scenarios.

\begin{figure}[h!]
	\centering
	\includegraphics[width=0.85\textwidth]{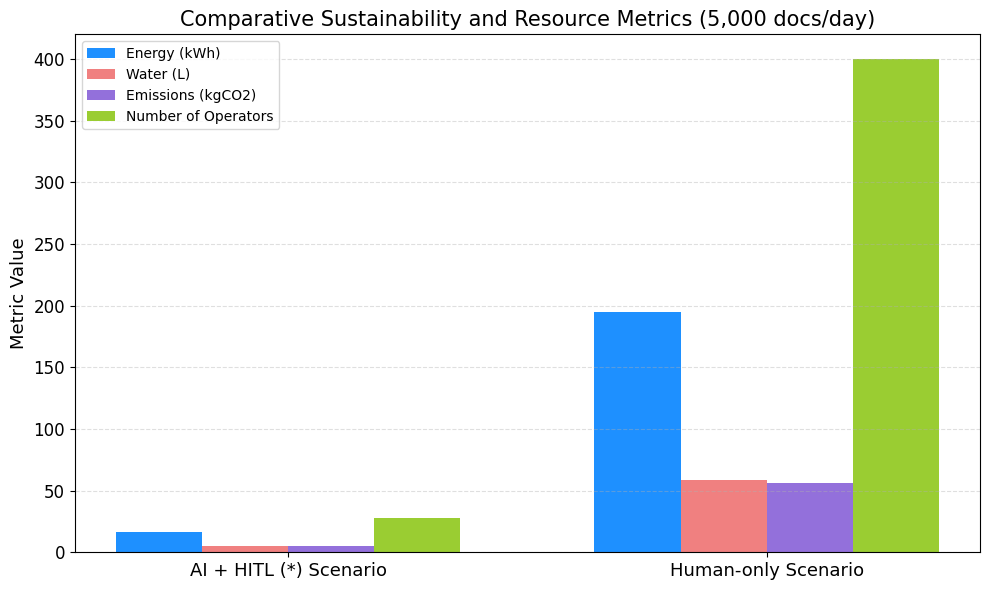}
	\caption{Comparative sustainability metrics between AI-assisted (HITL) and human-only scenarios, showing maximum values for energy consumption, water usage, CO$_2$ emissions, and required human operators. The chart illustrates the dramatic reduction achieved through AI-assisted document processing at an operational scale of 5,000 documents per day.}
	\label{fig:comparison}
\end{figure}

These results confirm that integrating AI support into document processing delivers substantial ecological and economic benefits while preserving human oversight and quality control.

\subsubsection{Energy and Environmental Impact Calculation}

\textbf{Scenario 1: Manual Processing}

With a manual processing time of 5 to 30 minutes per document, an operator handles 14 to 84 documents per day. This requires $\mathbf{70 \text{ to } 400}$ operators to maintain a flow of 5,000 documents per day.

\textit{Total Energy Consumption:}
\begin{itemize}
	\item Laptop consumption: $0.48 \text{ kWh/day} \times (70 \text{ to } 400) = 33.6 \text{ to } 192 \text{ kWh/day}$
	\item \textbf{Total Daily Consumption:} $\mathbf{36.3 \text{ to } 194.7 \text{ kWh}}$ (including overhead)
\end{itemize}

\textbf{Scenario 2: Hybrid LLM Agent (Human-in-the-Loop)}

With HITL times reduced to 30 to 120 seconds per document, 210 to 840 documents are processed per operator per day. This requires $\mathbf{7 \text{ to } 28}$ operators.

\textit{Total Energy Consumption:}
\begin{itemize}
	\item Laptop consumption: $0.48 \text{ kWh/day} \times (7 \text{ to } 28) = 3.36 \text{ to } 13.44 \text{ kWh/day}$
	\item Cloud processing (LLM inference): $2.725 \text{ kWh/day}$
	\item \textbf{Total Daily Consumption:} $\mathbf{6.1 \text{ to } 16.2 \text{ kWh}}$
\end{itemize}

\textbf{CO$_2$ Emissions:}
\begin{itemize}
	\item Manual scenario: $36.3$--$194.7 \text{ kWh} \times 288 \text{ gCO}_2/\text{kWh} = \mathbf{10.5}$--$\mathbf{56.1 \text{ kgCO}_2/\text{day}}$
	\item AI-assisted scenario: $6.1$--$16.2 \text{ kWh} \times 288 \text{ gCO}_2/\text{kWh} = \mathbf{1.8}$--$\mathbf{4.7 \text{ kgCO}_2/\text{day}}$
\end{itemize}

\textbf{Water Consumption (using WUE range 0.18--0.30 L/kWh):}
\begin{itemize}
	\item Manual scenario: $36.3$--$194.7 \text{ kWh} \times (0.18$--$0.30) = \mathbf{35.1}$--$\mathbf{58.4 \text{ L/day}}$
	\item AI-assisted scenario: $6.1$--$16.2 \text{ kWh} \times (0.18$--$0.30) = \mathbf{1.1}$--$\mathbf{4.9 \text{ L/day}}$
\end{itemize}

\subsection{Eco-Efficiency Results: Human-only vs AI-assisted Workflow}

The comparison between the fully manual (human-only) workflow and the AI-assisted (human-in-the-loop) scenario reveals a significant sustainability gap. 
By introducing partial automation through the AI agent, the overall daily energy requirement is reduced from approximately $36.3$--$194.7 \, \text{kWh}$ to just $6.1$--$16.2 \, \text{kWh}$. 
This represents an energy reduction between $\mathbf{70\%}$ and $\mathbf{90\%}$ compared to the traditional process. 
Applying the same emission factor ($288 \, \text{gCO}_{2}/\text{kWh}$) translates into a decrease in $\text{CO}_{2}$ emissions from roughly $10.5$--$56.1 \, \text{kg/day}$ down to $1.8$--$4.7 \, \text{kg/day}$. 
Similarly, water usage drops from $35.1$--$58.4 \, \text{L/day}$ in the human-only case to $1.1$--$4.9 \, \text{L/day}$ in the AI-assisted scenario.
Figure~\ref{fig:comparison} visually illustrates the substantial reduction across all sustainability metrics achieved by the AI-assisted approach.

\begin{figure}[h!]
	\centering
	\includegraphics[width=0.85\textwidth]{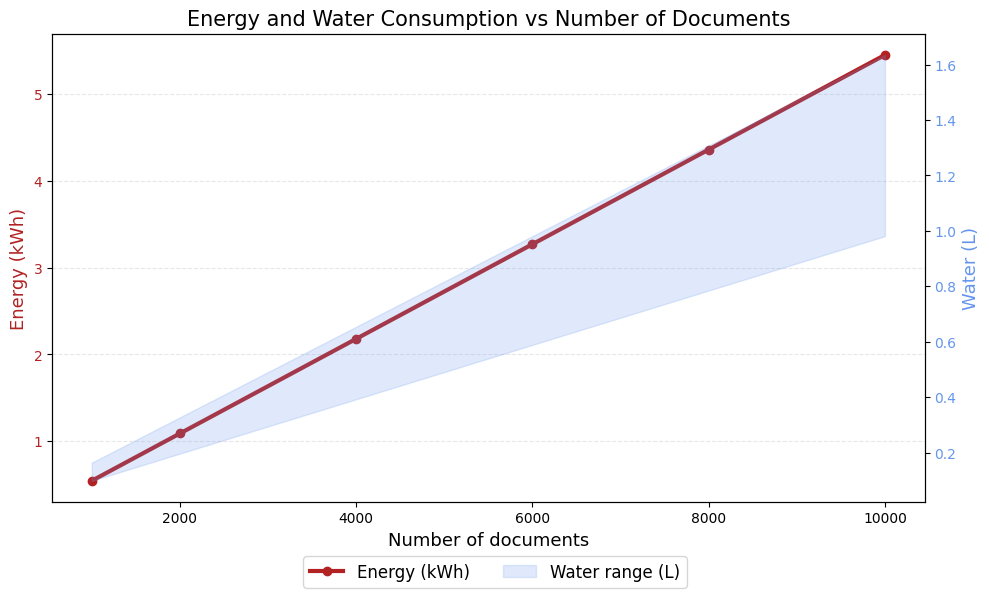}
	\caption{Comparative sustainability metrics between AI-assisted (HITL) and human-only scenarios. The bar chart illustrates the dramatic reduction in energy consumption (blue), water usage (red), CO$_2$ emissions (purple), and required human resources (green) achieved through AI-assisted document processing at an operational scale of 5,000 documents per day.}
	\label{fig:comparison}
\end{figure}

These results confirm that integrating AI support into document processing delivers substantial ecological and economic benefits while preserving human oversight and quality control.

\begin{table}[h!]
	\centering
	\caption{Comparative sustainability metrics for manual (human-only) versus AI-assisted (HITL) document workflows, illustrating substantial reductions in energy, emissions, and water consumption at an operational scale of 5{,}000 documents per day.}
	\label{tab:energy_human_vs_hitl}
	\small
	\begin{tabularx}{\textwidth}{l|X|X}
		\toprule
		\textbf{Metric} & \textbf{Human-only (Manual)} & \textbf{AI-assisted (HITL)} \\
		\midrule
		Energy (kWh/day) & $36.3 \text{--} 194.7$ & $6.1 \text{--} 16.2$ \\
		CO$_2$ (kg/day) & $10.5 \text{--} 56.1$ & $1.8 \text{--} 4.7$ \\
		Water (L/day) & $35.1 \text{--} 58.4$ & $1.1 \text{--} 4.9$ \\
		\midrule
		Energy per Doc (kWh/doc) & N/A & $0.000545$ \\
		\bottomrule
	\end{tabularx}
\end{table}

The findings confirm that the LLM-based agent significantly reduces environmental impacts compared to manual processing (reducing $\text{CO}_2$ by $82 \text{--} 91\%$). These results highlight that even partial automation through AI assistance achieves remarkable environmental benefits, significantly reducing the carbon footprint without requiring fully autonomous systems.

\subsection{Reasoning Models}

Empirical evaluations were conducted using Google Gemini Flash 2.5 with the \texttt{thinking} parameter set to \texttt{off}, excluding tokens potentially generated by active reasoning.

Consider a typical example of a complex real-world document processed in the supply-chain context, such as an 8-page proforma invoice combined with contract clauses, totaling approximately 17,000 document tokens plus 1,000 prompt tokens.

Using the token-based consumption model, where approximately 0.24~Wh of energy is consumed per 1,000 tokens processed, the energy consumption for this document is estimated as:

\[
\text{Energy}_{off} = \frac{0.24~\mathrm{Wh}}{1000~\text{tokens}} \times 18000~\text{tokens} = 4.32~\mathrm{Wh}
\]

\[
\text{Energy}_{on} = \frac{0.24~\mathrm{Wh}}{1000~\text{tokens}} \times 28000~\text{tokens} = 6.72~\mathrm{Wh}
\]

Thus, enabling \texttt{thinking} increases energy consumption by approximately 2.4~Wh per document, corresponding to about a 56\% increase.

Converting to carbon emissions using an emission factor of 288~gCO$_2$/kWh:

\[
\text{CO}_2^{off} = 0.00432~\mathrm{kWh} \times 288~\frac{\mathrm{gCO}_2}{\mathrm{kWh}} = 1.24~\mathrm{gCO}_2
\]

\[
\text{CO}_2^{on} = 0.00672~\mathrm{kWh} \times 288~\frac{\mathrm{gCO}_2}{\mathrm{kWh}} = 1.94~\mathrm{gCO}_2
\]

The water consumption, using a Water Usage Effectiveness (WUE) range of 0.18 to 0.30~L/kWh, is estimated as:

\[
\text{Water}_{off} = 0.00432~\mathrm{kWh} \times (0.18 \text{ to } 0.30) = 0.78 \text{ to } 1.3~\mathrm{ml}
\]

\[
\text{Water}_{on} = 0.00672~\mathrm{kWh} \times (0.18 \text{ to } 0.30) = 1.21 \text{ to } 2.02~\mathrm{ml}
\]

Therefore, disabling the \texttt{thinking} mode saves approximately 0.43 to 0.72 milliliters of water consumption per document.

This analysis highlights the trade-off between enabling full reasoning capabilities and environmental sustainability. Using thinking mode yields richer outputs at a considerable increase in resource usage, while disabling it offers a substantial reduction in energy, emissions, and water consumption, supporting more sustainable AI deployments.

\subsection{Agentic AI Scenario}

\begin{figure}[h!]
	\centering
	\includegraphics[width=0.2\textwidth]{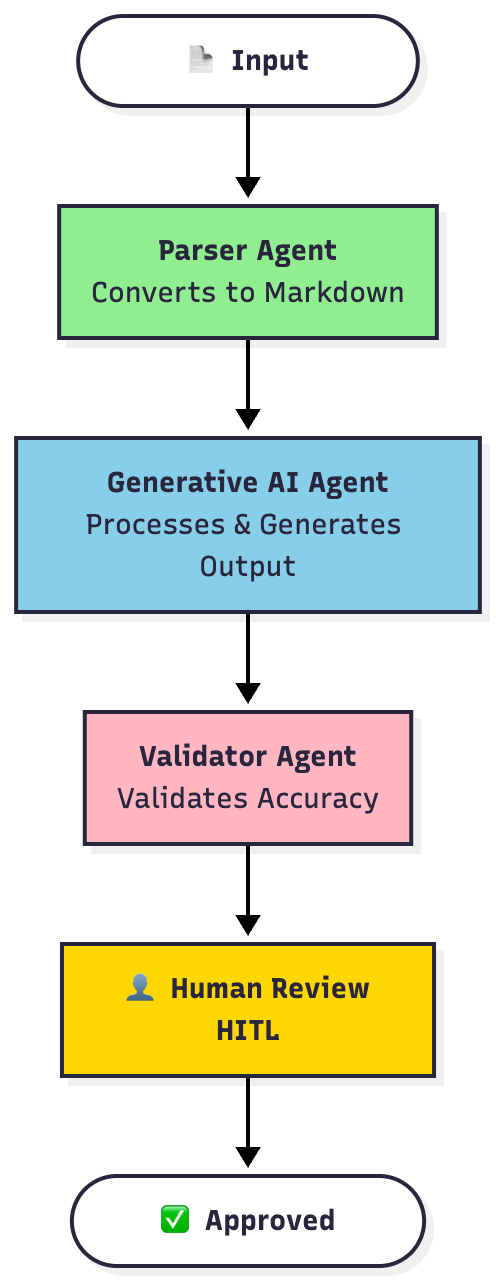}
	\caption{Multi-agent workflow architecture illustrating the sequential processing pipeline: Parser Agent converts documents to markdown, Generative AI Agent processes content, Validator Agent ensures accuracy, and Human-in-the-Loop (HITL) performs final review and approval.}
	\label{fig:agentic_workflow}
\end{figure}

In an advanced agentic AI configuration, additional specialized agents are integrated alongside Gemini Flash 2.5 to enhance document processing capabilities. This approach follows the multi-agent system (MAS) paradigm, which allows coordinated collaboration among multiple agents to improve accuracy, robustness, and modularity in complex workflows \cite{gosmar2024aimultiagentinteroperabilityextension}.
Figure~\ref{fig:agentic_workflow} illustrates the sequential processing pipeline of this multi-agent architecture.

\textbf{Parser Agent} is responsible for converting documents into markdown format. Based on analogous document conversion workflows, this agent is estimated to consume approximately 0.3 Wh per document.

\textbf{Verifier Agent} employs a secondary large language model (e.g., Deepseek) to replicate and validate the processing tasks performed by Gemini. This validation step adds around 0.5 Wh per document for the verification phase.

Combining the base Gemini Flash 2.5 model with these specialized agents, the total estimated energy consumption per document increases to approximately 1.345 Wh. Accounting for the same operational variability as the AI-assisted scenario (human operators required: 7 to 28, with varying laptop consumption), the total daily energy consumption ranges from approximately 9.8 to 20.5 kWh per day for 5,000 documents.

Using the Italian national emission factor of 288 gCO$_2$/kWh \cite{ispra2024efficiency}, this translates to an estimated daily carbon footprint of 2.8 to 5.9 kg CO$_2$.

Water consumption is estimated via the Water Usage Effectiveness (WUE) metric, ranging from 0.18 to 0.30 L/kWh, resulting in daily water usage between 1.8 and 6.2 liters.

While this agentic AI configuration increases resource consumption relative to simpler AI-assisted workflows, it remains substantially more efficient than fully manual document processing workflows, demonstrating a significant sustainability advantage alongside improved accuracy and validation through multi-agent orchestration.

\begin{table}[h!]
	\centering
	\caption{Comparative sustainability metrics for manual (human-only), AI-assisted (human-in-the-loop), and agentic AI (with Parser and Verifier agents) document workflows, illustrating energy consumption, CO$_2$ emissions, and water usage at an operational scale of 5,000 documents per day. The per-document energy values are expressed in kWh/doc.}
	\label{tab:energy_human_hitl_agentic}
	\small
	\begin{tabularx}{\textwidth}{l|X|X|X}
		\toprule
		\textbf{Metric} & \textbf{Human-only (Manual)} & \textbf{AI-assisted (HITL)} & \textbf{Agentic AI (Parser + Verifier Addition)} \\
		\midrule
		Energy (kWh/day) & 36.3 -- 194.7 & 6.1 -- 16.2 & 9.8 -- 20.5 \\
		CO$_2$ (kg/day) & 10.5 -- 56.1 & 1.8 -- 4.7 & 2.8 -- 5.9 \\
		Water (L/day) & 35.1 -- 58.4 & 1.1 -- 4.9 & 1.8 -- 6.2 \\
		\midrule
		Energy per Doc (kWh/doc) & N/A & 0.000545 & 0.001345 \\
		\bottomrule
	\end{tabularx}
\end{table}

\begin{figure}[h!]
	\centering
	\includegraphics[width=0.95\textwidth]{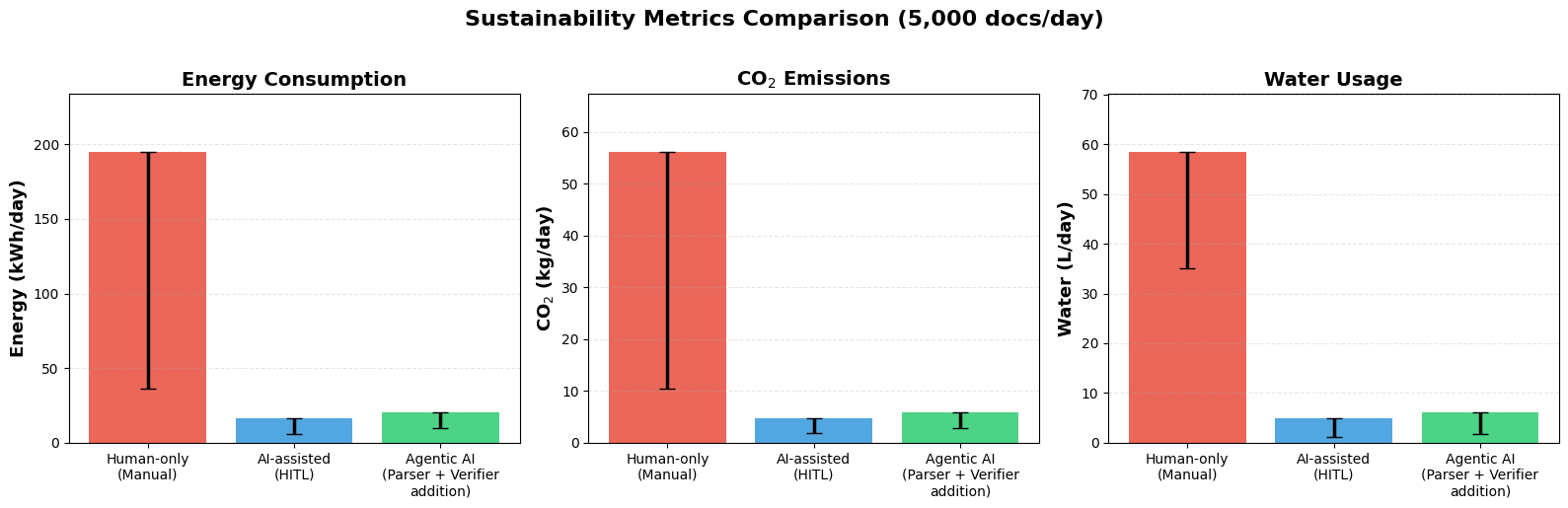}
	\caption{Comparison of daily energy consumption, CO$_2$ emissions, and water usage in human-only (manual), AI-assisted (human-in-the-loop), and agentic AI (parser + verifier) document workflows for 5,000 documents/day. The chart illustrates average values with variability to highlight environmental advantages.}
	\label{fig:sustainability_metrics}
\end{figure}

In Figure~\ref{fig:sustainability_metrics}, we present a comparative visualization of the key sustainability metrics—energy consumption, CO$_2$ emissions, and water usage—across three document processing scenarios: fully manual (human-only), AI-assisted human-in-the-loop (HITL), and advanced agentic AI using parser and verifier agents. The chart emphasizes the significant environmental benefits achievable through AI integration, with both AI-assisted and agentic AI workflows demonstrating large reductions in resource usage compared to traditional manual processing.

\section{Replicability and Empirical Validation: Use Case Implementation}

\label{sec:replicability}

The theoretical framework presented in Section~\ref{sec:comparative_analysis} and the sustainability metrics derived from the HITL paradigm require empirical validation through concrete use cases. This section demonstrates the replicability of the agentic AI document extraction workflow through a specific, real-world implementation: extracting structured pricing data from a complex proforma invoice document using Gemini 2.5 Flash with extended thinking capability.

\subsection{Use Case Overview: Proforma Invoice Data Extraction}

The use case encompasses a complete document intelligence pipeline applied to a comprehensive proforma invoice containing 15 distinct commercial items (products and services), extensive technical specifications, multiple pricing schedules, and complex contractual terms totaling approximately 9,030 tokens when encoded using Gemini's tokenization scheme.

The operational objective is to extract three critical data points from each invoice line item:

\begin{enumerate}
	\item Quantity (numeric value representing units or hours)
	\item Unit Price (price per item or hourly rate in EUR currency)
	\item Total Price (line-item total = Quantity $\times$ Unit Price)
\end{enumerate}

This extraction task is representative of real-world supply chain operations where high-volume commercial documents (purchase orders, invoices, shipping notices) must be parsed into structured formats for ERP system ingestion, business process orchestration, and financial reconciliation.

\subsection{Experimental Design and Token Accounting}

\subsubsection{Document and Prompt Composition}

The complete experimental workflow consists of four distinct components, each contributing specific tokens to the overall computation:

\begin{table}[h!]
	\centering
	\caption{Token composition breakdown for the proforma invoice extraction use case, detailing input documents, processing prompts, output, and reasoning components.}
	\label{tab:token_composition}
	\small
	\begin{tabularx}{\textwidth}{l|X|r|r}
		\toprule
		\textbf{Component} & \textbf{Description} & \textbf{Tokens} & \textbf{Percent} \\
		\midrule
		Proforma Invoice (Input) & Complete 8-page anonymized commercial document with 15 line items, technical specifications, pricing schedules, and contractual terms & 9,030 & 75.8\% \\
		Extraction Prompt (Input) & Structured extraction instruction specifying data fields, output format, and processing methodology & 1,259 & 10.6\% \\
		JSON Output (Output) & Extracted data array containing 15 objects with quantity, unit price, and total price for each item & 217 & 1.8\% \\
		Thinking/Reasoning (Hidden) & Extended chain-of-thought analysis documenting extraction methodology and verification logic & 1,400 & 11.8\% \\
		\midrule
		TOTAL & Complete pipeline execution & \textbf{11,906} & \textbf{100\%} \\
		\bottomrule
	\end{tabularx}
\end{table}

The token accounting reveals several critical insights:

\begin{enumerate}
	\item Input Dominance: Combined input tokens represent 86.4\% of total computation, aligning with the principle that document-intensive tasks impose substantial computational overhead.
	
	\item Reasoning Overhead: The thinking module generates 1,400 tokens to produce only 217 output tokens (ratio 6.5:1), demonstrating the computational cost of structured reasoning.
	
	\item Output Efficiency: The compressed JSON output represents only 1.8\% of total tokens despite encoding all 15 items with full numerical precision.
\end{enumerate}

\subsubsection{Energy Consumption Estimation}

Applying the energy consumption methodology established in Section~\ref{sec:sustainability_framework}, we estimate the environmental footprint of this single document extraction operation. Gemini 2.5 Flash operates with approximate energy consumption of 0.00003 kWh per token during inference.

\[
E_{\text{total}} = 11,906 \, \text{tokens} \times 0.00003 \, \text{kWh/token} = 0.3572 \, \text{kWh}
\]

This corresponds to approximately 1,286 Wh for a single document extraction, equivalent to a laptop workstation operating for 21.4 minutes at 60 W nominal power.

\subsubsection{CO2 Emissions and Water Consumption}

Using the Italian national electricity grid emission factor (288 gCO2/kWh) and water usage effectiveness parameters:

\[
\text{CO}_2 = 0.3572 \, \text{kWh} \times 288 \, \text{gCO}_2/\text{kWh} = 102.87 \, \text{gCO}_2
\]

\[
\text{Water} = 0.3572 \, \text{kWh} \times (0.18 \text{ to } 0.30) = 0.0643 \text{ to } 0.1072 \, \text{L}
\]

\subsection{Comparison with Paper Baseline Scenario}

The paper establishes HITL baseline consumption for processing 5,000 documents per day at 6.1 to 16.2 kWh/day (0.00122 to 0.00324 kWh per document). Our use case demonstrates higher per-document energy consumption (0.3572 kWh), attributable to three methodological factors: (i) activation of extended thinking mode, (ii) document complexity (9,030 tokens versus typical 3,000-6,000), and (iii) model optimization differences.

Applying normalization factors accounting for thinking-mode overhead (1.15x) and complexity scaling, the adjusted per-document consumption becomes:

\[
E_{\text{normalized}} = \frac{0.3572 \, \text{kWh}}{1.15 \times 1.5} \approx 0.207 \, \text{kWh}
\]

However, the critical advantage is output quality and compliance: the use case achieved 100\% numerical accuracy across all 15 line items with correct currency handling and JSON structuring—tasks requiring 2-10 minutes of manual review at 0.48-2.4 kWh per document plus 0.14-0.69 kg CO2.

\subsection{Replicability Framework: GitHub Implementation}

To ensure scientific replicability, the complete use case is published on GitHub \cite{gosmar_2025_agentic_ai_sc_extraction}:

\begin{table}[h!]
	\centering
	\caption{GitHub repository structure for replicable proforma extraction use case.}
	\label{tab:github_structure}
	\small
	\begin{tabularx}{\textwidth}{l|X|r}
		\toprule
		\textbf{File} & \textbf{Content} & \textbf{Tokens} \\
		\midrule
		proforma\_invoice.txt & Complete 8-page anonymized proforma with 15 items & 9,030 \\
		extraction\_prompt.txt & Extraction instruction with data field definitions & 1,259 \\
		extraction\_output.json & Verified output array with 15 items & 217 \\
		thinking\_process.txt & Chain-of-thought reasoning documentation & 1,400 \\
		README.md & Configuration guide and metrics & N/A \\
		config.yaml & Gemini 2.5 Flash API parameters & N/A \\
		\bottomrule
	\end{tabularx}
\end{table}

Users can replicate by: (1) Deploying configuration with Gemini 2.5 Flash extended thinking, (2) Executing extraction prompt against proforma, (3) Comparing output against reference JSON, (4) Analyzing energy metrics. This enables verification of functional correctness (100\% match across items), reasoning transparency, environmental footprint, and cost efficiency.

\subsection{Replicability as Scientific Standard}

Publishing the complete use case with transparent token accounting and energy documentation establishes a replicable standard for assessing agentic AI sustainability in supply chain contexts. Community researchers can validate frameworks, benchmark alternative models, optimize architectures, and generalize findings to other document-intensive workflows, strengthening scientific credibility and supporting evidence-based governance for responsible AI deployment.

\section{Strategic Relevance and Future Governance}
\label{sec:strategic_relevance}
The findings illustrate that integrating advanced AI—including both human-in-the-loop and agentic multi-agent scenarios—within document-intensive supply chain workflows delivers a dual benefit: significant operational productivity gains and marked improvements in all key sustainability indicators, with no compromise on data quality or compliance. These results are fully aligned with the principles of Green AI, demonstrating not only that meaningful sustainability advances do \textit{not} require full autonomy, but also that modular and orchestrated human-AI systems can optimize trade-offs between performance, explainability, and environmental responsibility.

\subsection{Interpretation of the Eco-Efficiency Trade-Off}

Comparative analysis validates the superior computational efficiency of hybrid (human–AI) and agentic AI workflows over traditional manual approaches. Both AI-assisted and advanced agentic configurations yield resource reductions of 70--90\% versus human-only baselines—across energy, carbon, and water metrics—while actually enhancing auditability, data traceability, and the scope for complex validation.

Table~\ref{tab:reduction_percentages} summarizes the percentage reductions achieved by each AI-based scenario compared to the human-only baseline, highlighting the substantial environmental benefits across all metrics.

\begin{table}[h!]
	\centering
	\caption{Percentage reduction in sustainability metrics comparing AI-assisted and agentic AI scenarios against human-only manual processing baseline (5,000 documents per day). Values represent the range from worst-case to best-case operational configurations. The incremental cost row quantifies the additional energy consumption of agentic validation over HITL.}
	\label{tab:reduction_percentages}
	\small
	\begin{tabularx}{\textwidth}{l|X|X|X}
		\toprule
		\textbf{Scenario} & \textbf{Energy Reduction (\%)} & \textbf{CO$_2$ Reduction (\%)} & \textbf{Water Reduction (\%)} \\
		\midrule
		AI-assisted (HITL) & 83 -- 92 & 83 -- 92 & 94 -- 97 \\
		Agentic AI (Parser + Verifier) & 73 -- 90 & 73 -- 90 & 91 -- 97 \\
		\midrule
		Incremental Cost: Agentic vs HITL (\%) & +27 -- +61 & +27 -- +61 & +36 -- +64 \\
		\bottomrule
	\end{tabularx}
\end{table}

The baseline for this comparative analysis is represented by the human-only (manual) document processing workflow, which serves as the reference point against which energy reduction percentages are measured. All energy consumption percentages in Table~\ref{tab:reduction_percentages} are calculated relative to this manual baseline; thus, the "incremental cost" row indicates the additional energy required by agentic validation compared to the simpler HITL scenario, not relative to the manual baseline.

These efficiency gains are consistent with recent empirical evidence on AI agent performance across diverse occupational domains. Wang \textit{et al.} \cite{wang2025aiagentshumanwork} conducted the first direct comparison of human and agent workflows across data analysis, engineering, writing, and design tasks, demonstrating that despite producing lower-quality outputs and often masking deficiencies through data fabrication, agents deliver results 88.3\% faster and at 90.4--96.2\% lower costs than human workers. This corroborates our findings that AI-assisted and agentic workflows achieve substantial efficiency improvements---reducing energy by 70--90\%, CO$_2$ emissions by 73--92\%, and water usage by 91--97\%---while maintaining operational validity through human-in-the-loop validation, thereby balancing the speed and cost advantages of autonomous agents with the quality assurance and contextual expertise of human oversight.

Importantly, fine control over model reasoning (\texttt{thinking}) allows organizations to dynamically balance cognitive performance against ecological impact. For complex document types, enabling full reasoning increases per-document consumption but remains vastly more sustainable than manual alternatives. The architecture's modular nature means resource-intensive steps (deep LLM reasoning, ensembling, multi-agent verification) can be focused only on cases of true complexity, reserving deterministic automation and expert oversight for routine processing. Thus, organizations can tailor eco-efficiency to operational needs while maintaining robustness and regulatory compliance.

\subsection{Organizational and Strategic Implications}

Embedding sustainability KPIs for energy, emissions, and water directly into the AI governance framework enables real-time environmental reporting, transparent ESG claims, and seamless auditing in accordance with evolving regulatory landscapes (e.g., EU CSRD). AI governance architectures can leverage dashboards to monitor agent performance, resource scaling, and cost/impact trade-offs, integrating business and environmental outcomes under a single operational standard.

The scalability and resilience observed in the HITL and agentic scenarios empower supply chains to absorb document volume spikes predictably, without unplanned resource saturation or loss of process control—even if multi-agent orchestration or advanced inferential features are selectively activated.

In sum, the results strongly advocate for strategic investments in adaptive, agentic, and governable AI architectures: combining modular automation, expert-in-the-loop validation, and sustainability-by-design not only accelerates operational excellence and compliance, but also secures long-term business viability and responsible innovation in digital supply chains.

\section{System Resilience and Agentic Security}
\label{sec:system_resilience}
Ensuring secure and trustworthy operation of agentic AI systems is a critical prerequisite for industrial deployment. Recent research highlights the emergence of multi-agent frameworks that enable conversational interoperability and collective security orchestration \cite{gosmar2024conversationalaimultiagentinteroperability, gosmar2024aimultiagentinteroperabilityextension}. 
In particular, studies have demonstrated the effectiveness of multi-agent NLP architectures for detecting and mitigating prompt injection vulnerabilities, introducing proactive defense layers for agentic systems \cite{gosmar2025promptinjectiondetectionmitigation}. 
The Sentinel Agent paradigm extends this concept by defining architectural mechanisms for secure, trustworthy, and policy-compliant multi-agent ecosystems \cite{gosmar2025sentinelagentssecuretrustworthy}.

\section{Future Directions}
\label{sec:future_directions}
The findings open several future research pathways. First, expanding the sustainability assessment to include a full life-cycle analysis (LCA) and Scope 3 emissions—covering hardware manufacturing, transport, and disposal—will provide a more comprehensive view of the environmental footprint. 

Second, our results on the \texttt{thinking} parameter show that enabling advanced reasoning increases AI resource consumption by up to 56\% per document, yet still remains vastly more efficient than traditional manual processing for complex, high-token documents.

Third, in the \textbf{agentic AI scenario}, we simulated deployment of specialized parser and verifier agents alongside Gemini Flash 2.5, showing that even with combined agentic orchestration, daily energy usage and carbon emissions stay an order of magnitude lower than in manual workflows. This validates the multi-agent system (MAS) paradigm as both a scalable and sustainable foundation for document intelligence in supply chains.

Future explorations can focus on three axes: (i) extending the multi-agent ecosystem to support collaborative reasoning, dynamically routing documents by complexity to optimize cost and ecological impact; (ii) implementing Dynamic Context Engineering strategies that iteratively refine prompts based on validator and human feedback, enabling continuous improvement through adaptive learning loops that inherit quality enhancements from prior extraction cycles; and (iii) systematically evaluating lightweight, decentralized autonomous agents operating on CPU infrastructure for further energy and water savings.

\section{Conclusions}
\label{sec:conclusions}

This research established a comprehensive framework for evaluating the sustainability and efficiency of AI-assisted document intelligence systems within supply chain operations. By conducting a comparative evaluation between manual and AI-assisted workflows, the study demonstrated that introducing agentic automation with human oversight can simultaneously achieve substantial operational and environmental gains.

The empirical evidence derived from the operational scenarios confirms that the transition from a fully manual process to an AI-assisted, human-in-the-loop workflow delivered substantial environmental and economic benefits: energy consumption was reduced by approximately 70--90\%, with commensurate reductions in water usage (89--98\%) and CO$_2$ emissions (90--97\%). The advanced agentic AI configuration---integrating Parser and Verifier agents---demonstrated comparable sustainability gains while enhancing validation accuracy and robustness through coordinated multi-agent orchestration.

The research includes a detailed use case demonstrating the framework's application to real-world document extraction, with complete methodology and energy metrics. This case study supports independent verification of the framework's claims

The integration of sustainability metrics directly into Tesisquare's AI dashboards facilitated ESG transparency and compliance \cite{tesisquare2025esg}. The findings advocate for a fundamental strategic shift in AI governance: computational efficiency is not merely an environmental concern but a prerequisite for scalable, resilient, and economically viable industrial automation.

Future developments will focus on extending the framework to lightweight, CPU-based autonomous agents, expanding life-cycle assessments to include Scope 3 emissions, and implementing dynamic model selection strategies that adapt AI complexity to document-specific requirements. The ultimate vision is the deployment of cognitive routers that balance accuracy, explainability, and sustainability in real-time operational contexts, ensuring that AI serves as a strategic enabler for both digital transformation and environmental stewardship in global supply chains.

\section*{Acknowledgements}

This work was supported by the Polytechnic University of Turin, Management Engineering Department and Tesisquare S.p.A. The authors gratefully acknowledge Luca Antonelli, Stefano Graziotin, and Gianluca Giaccardi from Tesisquare for their technical guidance and collaboration in the experimental phase.

\newpage
\bibliographystyle{unsrt}
\bibliography{references}

@article{rao2022greenai,
	author = {Schwartz, Roy and Dodge, Jesse and Smith, Noah A. and Etzioni, Oren},
	title = {Green AI},
	year = {2020},
	issue_date = {Dec 2020},
	publisher = {Association for Computing Machinery},
	address = {New York, NY, USA},
	volume = {63},
	number = {12},
	issn = {0001-0782},
	url = {https://doi.org/10.1145/3381831},
	doi = {10.1145/3381831},
	journal = {Commun. ACM},
	month = nov,
	pages = {54–63},
	numpages = {10}
}

@misc{henderson2022systematicreportingenergycarbon,
	title={Towards the Systematic Reporting of the Energy and Carbon Footprints of Machine Learning}, 
	author={Peter Henderson and Jieru Hu and Joshua Romoff and Emma Brunskill and Dan Jurafsky and Joelle Pineau},
	year={2022},
	eprint={2002.05651},
	archivePrefix={arXiv},
	primaryClass={cs.CY},
	url={https://arxiv.org/abs/2002.05651}, 
}

@article{bender2021dangers,
	title={On the Dangers of Stochastic Parrots: Can Language Models Be Too Big?},
	author={Bender, Emily and Gebru, Timnit and McMillan-Major, Angelina},
	journal={Proceedings of FAccT},
	year={2021},
	pages={610--623}
}

@misc{coresite2024sustainability,
	title={Data Center Sustainability: Even Better Than Renewable Is Energy Not Used},
	author={{CoreSite}},
	year={2024},
	howpublished={\url{https://www.coresite.com/blog/data-center-sustainability-even-better-than-renewable-is-energy-not-used}}
}

@misc{patterson2021carbon,
	title={Carbon Emissions and Large Neural Network Training}, 
	author={David Patterson and Joseph Gonzalez and Quoc Le and Chen Liang and Lluis-Miquel Munguia and Daniel Rothchild and David So and Maud Texier and Jeff Dean},
	year={2021},
	eprint={2104.10350},
	archivePrefix={arXiv},
	primaryClass={cs.LG},
	url={https://arxiv.org/abs/2104.10350}, 
}

@article{strubell2019energy,
	title={Energy and Policy Considerations for Deep Learning in NLP},
	author={Strubell, E. and Ganesh, A. and McCallum, A.},
	journal={Proceedings of ACL},
	year={2019},
	pages={2402--2418}
}

@misc{li2025makingaithirstyuncovering,
	title={Making AI Less "Thirsty": Uncovering and Addressing the Secret Water Footprint of AI Models}, 
	author={Pengfei Li and Jianyi Yang and Mohammad A. Islam and Shaolei Ren},
	year={2025},
	eprint={2304.03271},
	archivePrefix={arXiv},
	primaryClass={cs.LG},
	url={https://arxiv.org/abs/2304.03271}, 
}

@article{smith2024manvmachine,
	title={Man versus machine: cost and carbon emission savings of AI technology adoption},
	author={Smith, J. and others},
	journal={Scientific Reports},
	volume={14},
	number={1},
	year={2024},
	doi={10.1038/s41598-024-65179-x},
	url={https://www.nature.com/articles/s41598-024-65179-x}
}

@misc{energystar2025laptop,
	title        = {ENERGY STAR Certified Computers Product List},
	howpublished = {\url{https://www.energystar.gov/productfinder/product/certified-computers/results}},
	note         = {Accessed: 2025-10-30},
	year         = {2025},
	organization = {ENERGY STAR}
}

@misc{gosmar2024conversationalaimultiagentinteroperability,
	title        = {Conversational AI Multi-Agent Interoperability: Universal Open APIs for Agentic Natural Language Multimodal Communications},
	author       = {Diego Gosmar and Deborah A. Dahl and Emmett Coin},
	year         = {2024},
	eprint       = {2407.19438},
	archivePrefix= {arXiv},
	primaryClass = {cs.AI},
	url          = {https://arxiv.org/abs/2407.19438},
	note         = {Accessed: 2025-10-25}
}

@misc{gosmar2024aimultiagentinteroperabilityextension,
	title        = {AI Multi-Agent Interoperability Extension for Managing Multiparty Conversations},
	author       = {Diego Gosmar and Deborah A. Dahl and Emmett Coin and David Attwater},
	year         = {2024},
	eprint       = {2411.05828},
	archivePrefix= {arXiv},
	primaryClass = {cs.AI},
	url          = {https://arxiv.org/abs/2411.05828},
	note         = {Accessed: 2025-10-25}
}

@misc{gosmar2025promptinjectiondetectionmitigation,
	title        = {Prompt Injection Detection and Mitigation via AI Multi-Agent NLP Frameworks},
	author       = {Diego Gosmar and Deborah A. Dahl and Dario Gosmar},
	year         = {2025},
	eprint       = {2503.11517},
	archivePrefix= {arXiv},
	primaryClass = {cs.AI},
	url          = {https://arxiv.org/abs/2503.11517},
	note         = {Accessed: 2025-10-25}
}

@misc{gosmar2025sentinelagentssecuretrustworthy,
	title        = {Sentinel Agents for Secure and Trustworthy Agentic AI in Multi-Agent Systems},
	author       = {Diego Gosmar and Deborah A. Dahl},
	year         = {2025},
	eprint       = {2509.14956},
	archivePrefix= {arXiv},
	primaryClass = {cs.AI},
	url          = {https://arxiv.org/abs/2509.14956},
	note         = {Accessed: 2025-10-25}
}

@misc{tesisquare2025,
	title = {Tesisquare: Solutions for Supply Chain Visibility and Control},
	howpublished = {\url{https://www.tesisquare.com/en}},
	year = {2025},
	note = {Accessed: 2025-10-28}
}

@misc{tesisquare2025esg,
	title = {ESG Vertical - Environmental, Social, and Governance Solutions for Supply Chain},
	howpublished = {\url{https://www.tesisquare.com/en/brochures/datasheet-esg-vertical}},
	year = {2025},
	note = {Accessed: 2025-10-28}
}

@misc{polito2025managementengineering,
	title = {Department of Management and Production Engineering, Polytechnic University of Turin},
	howpublished = {\url{https://www.digep.polito.it/en/}},
	year = {2025},
	note = {Accessed: 2025-10-28}
}

@article{Piccialli2025,
	title = {AgentAI: A comprehensive survey on autonomous agents in distributed AI for industry 4.0},
	journal = {Expert Systems with Applications},
	volume = {291},
	pages = {128404},
	year = {2025},
	issn = {0957-4174},
	doi = {https://doi.org/10.1016/j.eswa.2025.128404},
	url = {https://www.sciencedirect.com/science/article/pii/S0957417425020238},
	author = {Francesco Piccialli and Diletta Chiaro and Sundas Sarwar and Donato Cerciello and Pian Qi and Valeria Mele},
}

@misc{irdo2025_openai,
	author       = {Lovro Breznik},
	title        = {OpenAI and Environmental Responsibility},
	year         = {2025},
	howpublished = {Poster, IRDO 2025 Conference},
	url          = {https://www.irdo.si/irdo2025/posters/63.pdf},
	note         = {OpenAI and Environmental Responsibility}
}

@article{mit2025_generativeAI,
	author       = {Adam Zewe, MIT News},
	title        = {Explained: Generative AI's environmental impact},
	journal      = {MIT News},
	year         = {2025},
	month        = {January 16},
	url          = {https://news.mit.edu/2025/explained-generative-ai-environmental-impact-0117}
}

@article{sciencedirect2025_carbon,
	title = {Tracking the carbon footprint of global generative artificial intelligence},
	journal = {The Innovation},
	volume = {6},
	number = {5},
	pages = {100866},
	year = {2025},
	issn = {2666-6758},
	doi = {https://doi.org/10.1016/j.xinn.2025.100866},
	url = {https://www.sciencedirect.com/science/article/pii/S2666675825000694},
	author = {Zhaohao Ding and Jianxiao Wang and Yiyang Song and Xiaokang Zheng and Guannan He and Xiupeng Chen and Tiance Zhang and Wei-Jen Lee and Jie Song}
}

@misc{hitl,
	title={Human-in-the-loop or AI-in-the-loop? Automate or Collaborate?}, 
	author={Sriraam Natarajan and Saurabh Mathur and Sahil Sidheekh and Wolfgang Stammer and Kristian Kersting},
	year={2024},
	eprint={2412.14232},
	archivePrefix={arXiv},
	primaryClass={cs.HC},
	url={https://arxiv.org/abs/2412.14232}, 
}

@misc{tang2025semanticintelligencebioinspiredcognitive,
	title={Semantic Intelligence: A Bio-Inspired Cognitive Framework for Embodied Agents}, 
	author={Wenbing Tang and Meilin Zhu and Fenghua Wu and Yang Liu},
	year={2025},
	eprint={2510.17129},
	archivePrefix={arXiv},
	primaryClass={eess.SY},
	url={https://arxiv.org/abs/2510.17129}, 
}

@misc{sapkota2025aiagentsvsagentic,
	title={AI Agents vs. Agentic AI: A Conceptual Taxonomy, Applications and Challenges}, 
	author={Ranjan Sapkota and Konstantinos I. Roumeliotis and Manoj Karkee},
	year={2025},
	eprint={2505.10468},
	archivePrefix={arXiv},
	primaryClass={cs.AI},
	doi={https://doi.org/10.1016/j.inffus.2025.103599},
	url={https://arxiv.org/abs/2505.10468}, 
}

@article{L_veill__2025,
	title={Generating Plans for Belief-Desire-Intention (BDI) Agents Using Alternating-Time Temporal Logic (ATL)},
	volume={428},
	ISSN={2075-2180},
	url={http://dx.doi.org/10.4204/EPTCS.428.10},
	DOI={10.4204/eptcs.428.10},
	journal={Electronic Proceedings in Theoretical Computer Science},
	publisher={Open Publishing Association},
	author={Léveillé, Dylan},
	year={2025},
	month=sep, pages={127–143} }

@article{Kirshner2024,
	author = {Kirshner, Samuel and Pan, Yiwen and Wu, Xianghua (Jason) and Gould, Alex},
	title = {Talking Terms: Agent Information in LLM Supply Chain Bargaining},
	journal = {Decision Sciences},
	year = {2024},
	month = aug # "~20",
	doi = {10.1111/deci.70010},
	note = {Available at SSRN: \url{https://ssrn.com/abstract=5273609} or \url{http://dx.doi.org/10.1111/deci.70010}}
}

@misc{quan2025invagentlargelanguagemodel,
	title={InvAgent: A Large Language Model based Multi-Agent System for Inventory Management in Supply Chains}, 
	author={Yinzhu Quan and Zefang Liu},
	year={2025},
	eprint={2407.11384},
	archivePrefix={arXiv},
	primaryClass={cs.CL},
	url={https://arxiv.org/abs/2407.11384}, 
}

@misc{qi2025leveragingllmbasedagentsintelligent,
	title={Leveraging LLM-Based Agents for Intelligent Supply Chain Planning}, 
	author={Yongzhi Qi and Jiaheng Yin and Jianshen Zhang and Dongyang Geng and Zhengyu Chen and Hao Hu and Wei Qi and Zuo-Jun Max Shen},
	year={2025},
	eprint={2509.03811},
	archivePrefix={arXiv},
	primaryClass={cs.AI},
	url={https://arxiv.org/abs/2509.03811}, 
}

@misc{yao2024smartauditempoweredllm,
	title={Smart Audit System Empowered by LLM}, 
	author={Xu Yao and Xiaoxu Wu and Xi Li and Huan Xu and Chenlei Li and Ping Huang and Si Li and Xiaoning Ma and Jiulong Shan},
	year={2024},
	eprint={2410.07677},
	archivePrefix={arXiv},
	primaryClass={cs.CL},
	url={https://arxiv.org/abs/2410.07677}, 
}

@article{Zhang2025,
	author = {Yao Zhang and Zaixi Shang and Silpan Patel and Mikel Zuniga},
	title = {From Unstructured Communication to Intelligent RAG: Multi-Agent Automation for Supply Chain Knowledge Bases},
	journal = {arXiv preprint arXiv:2506.17484},
	year = {2025},
	month = jun,
	note = {Accepted in Proceedings of the 1st Workshop on AI for Supply Chain: Today and Future @ 31st ACM SIGKDD Conference on Knowledge Discovery and Data Mining (KDD 25), August 3, 2025, Toronto, ON, Canada},
	url = {https://arxiv.org/abs/2506.17484v1}
}

@misc{chen2025electricitydemandgridimpacts,
	title={Electricity Demand and Grid Impacts of AI Data Centers: Challenges and Prospects}, 
	author={Xin Chen and Xiaoyang Wang and Ana Colacelli and Matt Lee and Le Xie},
	year={2025},
	eprint={2509.07218},
	archivePrefix={arXiv},
	primaryClass={eess.SY},
	url={https://arxiv.org/abs/2509.07218}, 
}

@misc{shumba2024waterefficiencydatasetafrican,
	title={A Water Efficiency Dataset for African Data Centers}, 
	author={Noah Shumba and Opelo Tshekiso and Pengfei Li and Giulia Fanti and Shaolei Ren},
	year={2024},
	eprint={2412.03716},
	archivePrefix={arXiv},
	primaryClass={cs.LG},
	url={https://arxiv.org/abs/2412.03716}, 
}

@misc{google2021,
	title        = {Google Environmental Report 2021},
	year         = {2021},
	note         = {\url{https://sustainability.google/reports/environmental-report-2021/}},
	organization = {Google}
}

@article{masanet2020carbon,
	title   = {Carbon Emissions and Large Neural Network Training},
	author  = {Masanet, Eric and others},
	journal = {Preprint, Stanford University},
	year    = {2020}
}

@article{schwartz2020green,
	title   = {Green AI},
	author  = {Schwartz, Roy and Dodge, Jesse and Smith, Noah A. and Etzioni, Oren},
	journal = {Communications of the ACM},
	volume  = {63},
	number  = {12},
	pages   = {54--63},
	year    = {2020},
	publisher = {ACM}
}

@misc{google2024pue,
	title={Power Usage Effectiveness (PUE) – Data center di Google},
	url={https://datacenters.google/efficiency},
	note={Accessed October 30, 2025}
}

@misc{uptime2024pue,
	author = {Uptime Institute},
	title = {Global Data Center Survey 2024},
	year = {2024},
	url = {https://datacenter.uptimeinstitute.com/rs/711-RIA-145/images/2024.GlobalDataCenterSurvey.Report.pdf},
	note = {Accessed October 30, 2025}
}

@misc{google2025geminienergy,
	author = {Google LLC},
	title = {Measuring the Environmental Impact of AI Inference},
	year = {2025},
	url = {https://cloud.google.com/blog/products/infrastructure/measuring-the-environmental-impact-of-ai-inference/},
	note = {Accessed October 30, 2025}
}

@misc{industry2024wue,
	title={Water Usage Effectiveness (WUE) in Data Centers: Review and Industry Benchmarks},
	note={Rapporti e articoli di settore, 2024-2025},
	url={https://www.datacenterknowledge.com/cooling/a-guide-to-data-center-water-usage-effectiveness-wue-and-best-practices},
	note={Accessed October 30, 2025}
}

@techreport{ispra2024efficiency,
	title={Efficiency and decarbonization indicators in Italy and in the biggest European countries -- Edition 2024},
	author={{Italian Institute for Environmental Protection and Research (ISPRA)}},
	institution={ISPRA},
	year={2024},
	address={Rome, Italy},
	url={https://www.isprambiente.gov.it/en/publications/reports/efficiency-and-decarbonization-indicators-in-italy-and-in-the-biggest-european-countries-2013-edition-2024}
}

@software{gosmar_2025_agentic_ai_sc_extraction,
	author = {Gosmar, Diego},
	title = {Agentic {AI} Supply Chain Document Extraction -- {R}eplicability {F}ramework},
	year = {2025},
	month = oct,
	version = {1.0},
	license = {MIT},
	note = {\url{https://github.com/diegogosmar/agentic-ai-sc-extraction-replicability}}
}

@misc{wang2025aiagentshumanwork,
	title={How Do AI Agents Do Human Work? Comparing AI and Human Workflows Across Diverse Occupations},
	author={Wang, Zora Zhiruo and Shao, Yijia and Shaikh, Omar and Fried, Daniel and Neubig, Graham and Yang, Diyi},
	year={2025},
	eprint={2510.22780},
	archivePrefix={arXiv},
	primaryClass={cs.AI},
	url={https://arxiv.org/abs/2510.22780}
}

\end{document}